\documentclass[runningheads]{llncs}
\usepackage{caption}
\usepackage{comment}
\usepackage[colorlinks=true, citecolor=blue, linkcolor=black, urlcolor=blue]{hyperref}
\usepackage[T1]{fontenc}
\usepackage{graphicx}

\usepackage{bbding}
\begin{document}
\title{Assisted Spatial Cognition Through Vision-Language Models}
\subtitle{A Navigation Support System for Visually Impaired and Neuro-divergent Users}
%
\author{Hamza Riaz\inst{1,2}\orcidID{0000-0001-6339-6194} \and
Jaime B. Fernandez\inst{1,2}\orcidID{0000-0001-9774-3879}\Envelope \and
Ian Mills\inst{3}\orcidID{0000-0003-3984-0755} \and
Daniel Hickey\inst{3}\orcidID{0000-0002-5339-992X} \and
Frances Cleary\inst{3}\orcidID{0000-0003-1823-0157} \and
Muhammad Intizar Ali\inst{1,2}\orcidID{0000-0002-0674-2131}}
\authorrunning{H. Riaz et al.}
%
\institute{School of Electronic Engineering, Dublin City University, Dublin, Ireland \and
Insight Research Ireland Centre for Data Analytics, Dublin, Ireland \and South East Technological University, Waterford, Ireland\\
\email{ hamza.riaz@dcu.ie, jaimeboanerjes.fernandezroblero@dcu.ie, ian.mills@waltoninstitute.ie, daniel.hickey@waltoninstitute.ie, frances.cleary@waltoninstitute.ie, ali.intizar@dcu.ie}\\
}
%

%
%

%
\maketitle              
\begin{abstract}
Multimodal Artificial Intelligence (AI), powered by Large Language Models (LLMs) and Vision-Language Models (VLMs), is transforming assistive technologies by enabling simultaneous processing of visual and textual data. This advancement holds significant promise for over 43 million visually impaired and neuro-divergent individuals worldwide who face persistent challenges in navigating indoor and outdoor environments due to limited spatial awareness and insufficient environmental cues. Existing navigation aids often lack comprehensive 3D scene understanding, relying on constrained route-based strategies that hinder user autonomy. In this paper, we introduce a novel end-to-end framework that integrates LLMs, VLMs and digital twin technologies to deliver a spatially cognitive navigation support for visually impaired and neuro-divergent users. Our system captures video input via standard mobile phone cameras, and employs SLAM3R to generate dense 3D point clouds from monocular RGB sequences in real-time. Our custom post-processing algorithm ensures accurate point cloud alignment across multiple viewpoints without requiring predefined reference points. This enhances the capabilities of SpatialLM to produce structured 3D representations, including architectural elements and oriented object bounding boxes. The enriched spatial data is then processed by a locally deployed LLM, which interprets 3D contexts to generate detailed scene descriptions and precise distance measurements between users and surrounding objects. We evaluated our approach across diverse video scenarios (30 seconds to 2 minutes) featuring various perspectives (e.g. frontal, $180^\circ$, $360^\circ$,) and looped walking views)  captured in multiple environments. The evaluation results demonstrate consistent accuracy in 3D scene interpretation and object localisation, underscoring the potential of our system as a transformative assistive navigation solution that combines advanced visual perception with spatial reasoning.

\end{abstract}

\section{Introduction}
Navigation and spatial awareness represent fundamental challenges for individuals with visual and cognitive impairments worldwide. According to the World Health Organization, over 2.2 billion people experience near or distance vision impairment, including 43 million living with blindness and 295 million with moderate-to-severe visual impairment~\cite{who2023blindness,orbis2021global}. Approximately 15-20\% of the global population is neuro-divergent, including individuals with autism spectrum disorders, ADHD, dyslexia, and other cognitive conditions that impact spatial reasoning and navigation capabilities~\cite{zurich2022neurodivergent}. These populations often encounter significant barriers when navigating complex indoor and outdoor environments, leading to reduced mobility, limited autonomy, and external assistance for spatial orientation and obstacle avoidance. 

Traditional assistive technologies predominantly rely on 2D image processing and scene analysis, which fundamentally limits their ability to provide accurate spatial information critical for safe navigation. Two-dimensional perspectives do not capture the essential depth relationships, precise object positioning, and volumetric scene structure necessary for effective wayfinding~\cite{liu2023open}. Visually impaired and neuro-divergent users require explicit distance measurements, spatial relationships between objects, and three-dimensional object localisation to make informed navigation decisions~\cite{MASHIATA2022100265}. Conventional 2D-based navigation systems lack the capacity to accurately convey depth information, such as distinguishing whether an obstacle is 2 meters or 10 meters away. Moreover, they are insufficient for representing the spatial complexity of environments, such as hospitals, university campuses, or transportation hubs, where precise 3D scene understanding is essential for enabling independent and confident mobility 

Recent advances in computer vision (CV), machine learning (ML), and AI have demonstrated significant progress in 3D scene understanding for assistive applications. Traditional approaches have employed geometric methods, stereo vision, and depth sensors combined with classical machine learning techniques for indoor navigation~\cite{10.1145/3733155.3734895}. Contemporary research has increasingly leveraged deep learning architectures, including convolutional neural networks for depth estimation, graph neural networks for spatial relationship modelling, and transformer-based approaches for scene graph generation~\cite{qi2025gpt4scene}. Most recently, the emergence of LLMs and VLMs has opened unprecedented opportunities for multimodal scene understanding, enabling systems to process both visual and textual information simultaneously for comprehensive spatial intelligence~\cite{qi2025gpt4scene}. 

This paper presents a novel end-to-end framework that addresses the limitations of existing assistive technologies by integrating state-of-the-art 3D reconstruction, spatial understanding, and natural language processing (NLP). Our approach uses SLAM3R for dense 3D point cloud generation from smartphone camera footage, followed by sophisticated point cloud alignment algorithms which we have developed from scratch to ensure spatial coherence across temporal sequences~\cite{liu2025slam3r,murai2025mast3r}. The aligned 3D representations are subsequently processed through SpatialLM, a specialised vision-language model trained for structured indoor spatial understanding, which generates semantically rich 3D scene descriptions with precise object localisation and bounding box predictions~\cite{mao2025spatiallm}. After this step, a local LLM GPT-OSS-20B model is deployed, which then synthesises this 3D spatial information to provide comprehensive, natural language descriptions of the environment, including accurate distance measurements and spatial relationships between the user and surrounding objects and obstacles. In this way, users can extract better spatial reasoning abilities from the models using prompt engineering. Figure \ref{figure1}  explains the overall data collection pipeline and Figure \ref{figure2} expresses the methodology of the proposed solution. Furthermore, the results section explains different visual outputs of the system, e.g. bird-eye-view in Figure \ref{figure4}, the overall results of different indoor environments in terms of 3D bounding boxes, and structural layout detection in Figure \ref{figure3}. Figure \ref{figure5} explains the 3D scene descriptions and the spatial reasoning (question-answers results) from GPT-OSS-20B.


The remainder of this paper is organised as follows: Section \ref{background} provides a comprehensive background 
 and section \ref{data_collection} details our systematic data collection. Section \ref{methodology} presents the complete methodology of our proposed integrated system, with video capture, 3D reconstruction, SpatialLM inference, and GPT-OSS-20B NLP. Section \ref{Use_cases_results} discusses our experiments across diverse indoor environments and also provides bird-eye view (BEV) visualisations and scene descriptions in natural language as generated by our locally deployed LLM. Finally, section \ref{conclusion_future_research} concludes with a summary of contributions and outlines future research directions for advancing assistive navigation technologies.

\section{Background} \label{background}

\subsection{Assistive Technologies for Visually Impaired and Neurodivergent Users}

Recent advances in artificial intelligence, particularly in computer vision and multimodal LLMs, have spawned revolutionary assistive technologies for people with visual and cognitive impairments~\cite{gamage2025vision}. The integration of AI into assistive technologies demonstrates transformative potential across healthcare, education, and daily living environments, with AI-powered devices supporting elderly autonomy through smart wheelchairs and exoskeletons, while AI-driven tools enhance social interaction for individuals with autism spectrum disorders. Moreover, systems like Microsoft's Seeing AI demonstrate capabilities in reading text, identifying objects, and describing scenes through computer vision and NLP~\cite{microsoft2019seeing}. These solutions remain constrained by their 2D perspective and cannot accurately convey precise spatial relationships and 3D environmental structure necessary for safe navigation in complex indoor and outdoor environments~\cite{10.1007/978-3-031-05039-8_12}. 


\subsection{SLAM-based 3D Scene Reconstruction}

Camera-based Simultaneous Localisation and Mapping (SLAM) has emerged as a critical technology for real-time 3D scene reconstruction, with several notable research contributions advancing the field. MASt3R-SLAM presents a real-time monocular dense SLAM system designed from MASt3R, a two-view 3D reconstruction works that explains robust performance on "in-the-wild" video sequences without assumptions on fixed camera models~\cite{murai2025mast3r}. FlashSLAM leverages 3D gaussian splatting for efficient and robust 3D scene reconstruction and addresses limitations of gradient descent-based optimisation through fast vision-based camera tracking with pretrained feature matching models and achieves 90\% reduction in tracking time compared to SplaTAM 
while maintaining superior accuracy in sparse view settings. 
DeepFusion integrates convolutional neural network outputs with semi-dense multiview stereo algorithms to produce fully dense depth maps for keyframes with metric scale which illustrates real-time dense reconstruction capabilities on GPU hardware.


\subsection{Vision-Language Models for 3D Scene Understanding}

VLMs have recently displayed remarkable potential in 3D spatial reasoning and scene understanding, with several groundbreaking approaches advancing the field. LEO represents an embodied multi-modal generalist agent that excels in perceiving, grounding, reasoning, planning, and acting in 3D environments through object-centric point-cloud processing and ego-view observations~\cite{huang2024embodied}. VLM-3R introduces a unified framework that incorporates 3D reconstructive instruction tuning, processing monocular video frames through geometry encoders to derive implicit 3D tokens representing spatial understanding, leveraging over 200K curated 3D reconstructive instruction tuning question-answer pairs~\cite{fan2025vlm}. Chat-Scene bridges 3D scenes and LLM through object identifiers and object-centric representations, decomposing input 3D scenes into object proposals with unique identifier tokens for efficient referencing and grounding~\cite{huang2024chat}. GPT4Scene introduces a novel visual prompting paradigm that enhances 3D spatial understanding by establishing global-local relationships through Bird's Eye View images and consistent object IDs, significantly improving zero-shot 3D comprehension tasks~\cite{qi2025gpt4scene}.

\section{Data Collection and Preparation} \label{data_collection}

To effectively fine-tune and deploy SpatialLM for our targeted 3D understanding tasks, we established a systematic data collection and preparation process, ensuring high-quality, appropriately orientated, and scaled point clouds as input \cite{mao2025spatiallm}.  The method starts with video acquisition on an iPhone 16 or android with a wide-angle lens, which captures steady, low-blur footage of inside areas, including college buildings, labs, hospitals, workplaces, train station, and shopping centres.  The videos were filmed using best practices from SLAM3R demos, such as smooth camera motions, constant pacing, and complete coverage of the space to minimise gaps in reconstruction \cite{liu2025slam3r}. Shaky or blurry footage can inject noise and artefacts into the resultant point clouds, reducing downstream model performance in tasks like layout detection and object detection \cite{6385773}. Therefore, frame stability and consistency are critical. To produce dense 3D point clouds from these monocular videos, we used SLAM3R, a cutting-edge simultaneous localisation and mapping (SLAM) system with real-time processing abilities \cite{liu2025slam3r}. 

\begin{figure*}[htbp]
\centering
\includegraphics[width=0.9\textwidth]{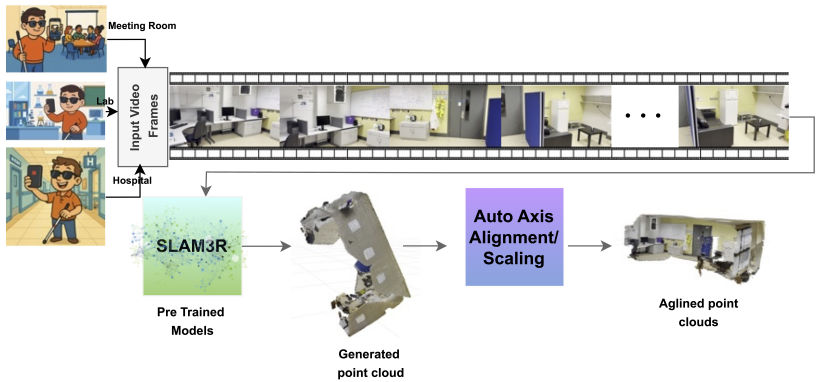} %
\caption{The diagram represents the process involve to produce a clean and high-quality point cloud using mobile camera, pre-trained model for SLAM applications, post-processing for alignment and scaling of data.}
\label{figure1}
\end{figure*}

Intermediate processing steps are necessary after point cloud generation to align and scale the data. SpatialLM requires input point clouds to be axis-aligned with the z-axis pointing upwards, adhering to the ScanNet orientation convention that transforms scans to z-up alignment and aligns walls to x-y planes \cite{dai2017scannet}. Improper orientation can result in poor performance in structured scene modelling, including inaccurate wall recognition and bounding box predictions.  
The script loads the point cloud from various formats (e.g., PLY, PCD, XYZ), optionally removes statistical outliers using Open3D's function with configurable neighbours and standard deviation ratio to eliminate noise from reconstruction artefacts, which is critical for maintaining clean inputs and preventing hallucinations in SpatialLM's autoregressive generation \cite{mao2025spatiallm}.  Next, it fixes the z-axis orientation by applying a transformation matrix that converts y-up to z-up conventions. Subsequent steps in the script correct minor floor tilts by estimating and rotating around the x-axis to level the ground plane. This is crucial for preventing skewed spatial representations that could cascade into errors. Finally, the script calculates a scale factor based on estimated room height (e.g., 2.5 m) and applies it evenly, standardising metric units to meet real-world scales necessary for embodied activities like distance-based reasoning ~\cite{mao2025spatiallm}. 
To improve clarity, the Figure \ref{figure1} illustrates the pipeline, such as multi-panel diagram showing raw video frames, pre trained model, generated point cloud before processing (with visible misalignments), after z-axis fix and tilt correction, and scaled final input with metric annotations \cite{zhou2018open3d}. 

\section{Methodology} \label{methodology}
The methodology encompasses four primary stages: video capture and preprocessing, SLAM3R-based 3D reconstruction, SpatialLM inference for structured scene understanding, and GPT-OSS-20B integration for natural language scene description and spatial reasoning.

\subsection{Video Capture and Preprocessing Pipeline}

Our system initiates with standard smartphone camera video capture, recording short sequences ranging from 30 seconds to 2 minutes in both indoor and outdoor environments. The preprocessing module converts raw video streams into suitable frame formats optimised for 3D reconstruction, including frame extraction at consistent temporal intervals, resolution standardisation, and color space normalisation to ensure compatibility with downstream computer vision models. 

\subsection{SLAM3R-based Point Cloud Generation: Aligned \& Scaled Point Clouds}
In this way, after getting the consecutive frames from the videos, we utilise these frames to create 3D point clouds of the input environments by using pre-trained weights of SLAM3R model. This approach gives 3D scene understanding by leveraging ubiquitous hardware rather than specialised depth sensors or stereo camera systems. The approach assumes access to high-quality, axis-aligned, metric-scale point clouds, as detailed in the data preparation section \ref{data_collection}. Specifically: \textbf{in Z-up orientation} the point cloud follows ScanNet's standard, with z as the up axis and walls aligned with the x-y planes. In the same way, \textbf{Alignment} coordinates the frame and makes sure it is aligned with the dominating room axes to eliminate geometric ambiguity. The \textbf{Metric scale} converts the scenes in meters to aid with distance-aware thinking and planning.

 \subsection{SpatialLM Inference for Structured Scene Understanding}
We chose SpatialLM as the primary scene-structuring model because of its ability to transform unstructured 3D geometry into structured indoor "code," which provides architectural primitives such as walls, doors, windows, and floor/ceiling planes. It also produces orientated 3D boxes that provide category names, locations, sizes, and yaw (with optional complete rotation where possible).

\paragraph{Structured Output Generation}
SpatialLM produces a text file that encodes a structured indoor model as a series of written entries, each including semantic class and 3D geometry.  The format is given in the SpatialLM paper \cite{mao2025spatiallm}, specifically, object class, position, size, and orientation. In practice, the file contains entries for 

\begin{itemize}
    \item \textbf{Architectural Elements}: Wall planes, door and window apertures, floor and ceiling boundaries with precise geometric parameters
    \item \textbf{3D Object Detection}: Oriented bounding boxes with semantic categories, precise 3D positions (x, y, z), dimensions (length, width, height), and orientation angles
    \item \textbf{Layout Topology}: Spatial relationships between architectural elements, room segmentation, and connectivity graphs representing navigable pathways
    \item \textbf{Semantic Classifications}: Object categories and functional annotations that enable high-level scene understanding.
\end{itemize}

\subsection{GPT-OSS-20b Integration for NLP}

The structured SpatialLM outputs are processed through a locally deployed GPT-OSS-20b model to generate comprehensive, human-interpretable scene descriptions and spatial reasoning capabilities. We selected GPT-OSS-20b for its superior deployment efficiency, achieving 31.8\% higher decode throughput and maintains high-quality natural language generation suitable for real-time assistive applications.

\subsubsection{3D Scene Description Generation}

Our scene description module synthesises SpatialLM's structured outputs into coherent, navigation-relevant descriptions like \textbf{Hierarchical Spatial Description}, \textbf{Distance-Aware Narration}, \textbf{Directional Guidance} like Spatial relationship descriptions using natural language (e.g., "chair 2.3 meters to your left," "doorway 4.5 meters ahead", and \textbf{Saliency-Based Prioritisation}, which is indicated in the results of Figures \ref{figure3}, \ref{figure4}, and \ref{figure5}. Similarly, for spatial reasoning and 3D question answering, the system enables interactive spatial reasoning through natural language queries such as \textbf{Geometric Relationship Queries}, \textbf{Object Localisation}, \textbf{Navigation Assistance}, and \textbf{Contextual Awareness}

\begin{figure*}[htbp]
\centering
\includegraphics[width=0.9\textwidth]{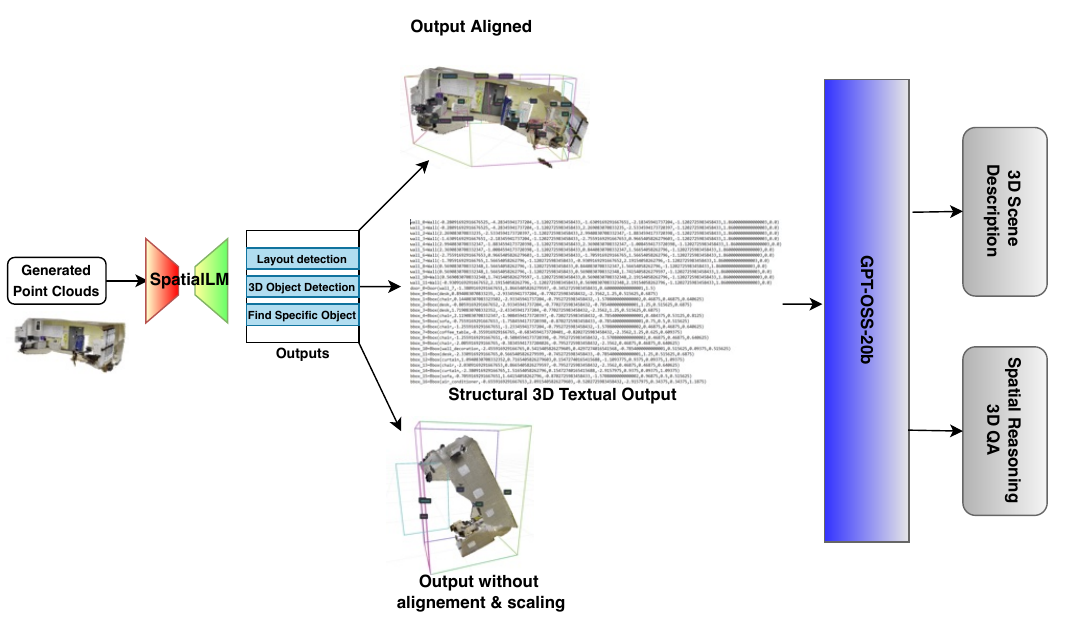} %
\caption{Complete pipeline architecture of our proposed assistive navigation system.} 
\label{figure2}
\end{figure*}

\section{Use Cases \& Results} \label{Use_cases_results}

Potential users for outdoor applications to travel from point A to point B can definitely use Google Maps; however, for indoor navigation visually impaired people still struggle to get precise navigation plans. Therefore, we think that to give complete precise perspective to visually impaired users, we need to explore 3D understanding-based solutions. Hence, we have tested our proposed solution in different indoor environments like laboratory, conference meeting room, and a big architectural building. As we only had access to laboratory and big architectural building environments, therefore, to test the robustness of the system, we collected video data under different conditions like with front view only, with $180^\circ$ view, with $360^\circ$ view, and by walking to the place in a loop. We tested generation of point clouds and results from the SpatialLM model on videos of different lengths including 30 seconds, 60 seconds, and 120 seconds.

\subsection{Testing Environments} 
\begin{figure*}[htbp]
\centering
\includegraphics[width=0.9\textwidth]{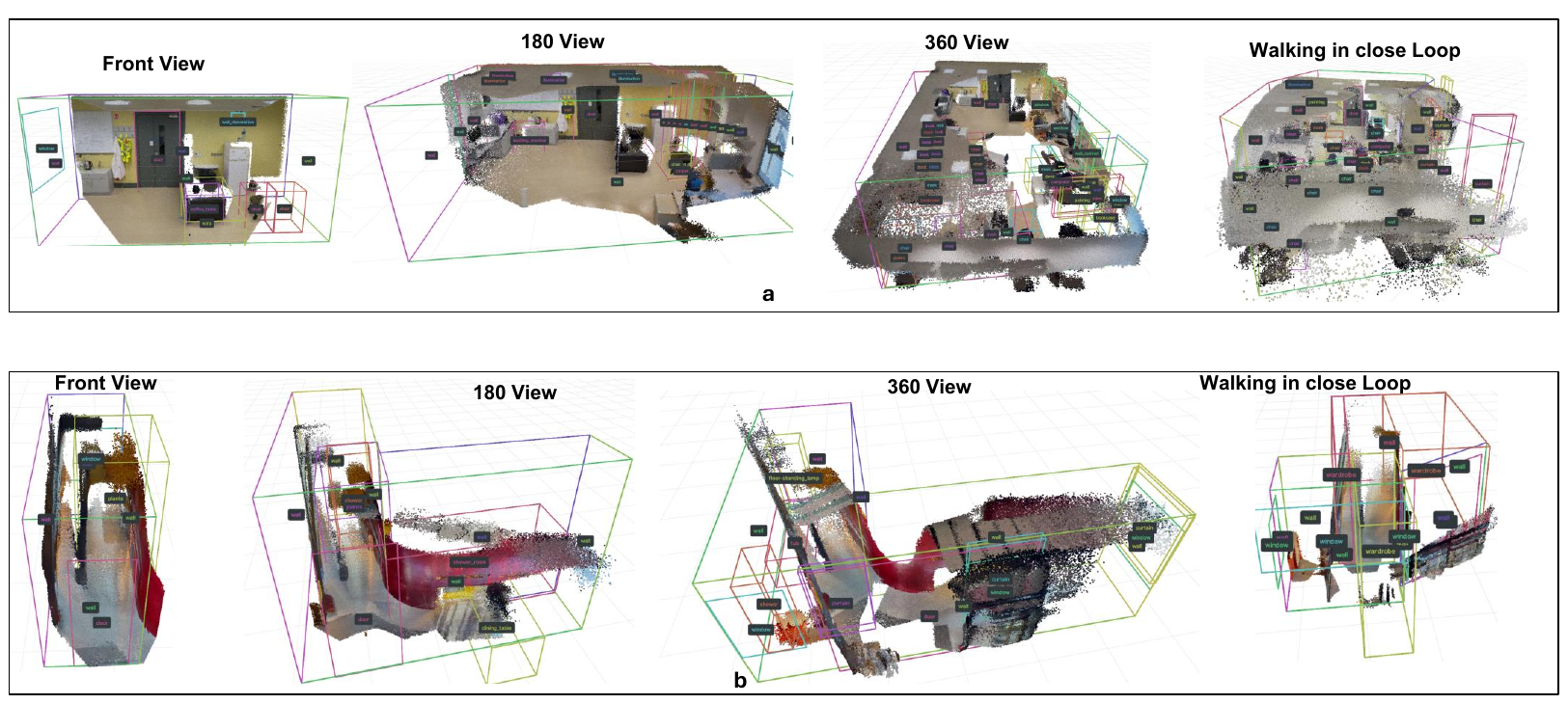} %
\caption{Figure represents visual results on generated point-clouds from the videos of our environments. (a) part of the Figure indicates results from front, $180^\circ$, $360^\circ$, and walking views of the laboratory. Moreover, (b) describes the big building architectural results.}
\label{figure3}
\end{figure*}

\subsubsection{Laboratory}
The laboratory environment represents a controlled indoor setting typical of academic and research workspaces, featuring structured layouts with desks, chairs, computer equipment, whiteboards, and storage areas. We captured a total of 12 videos within this laboratory environment designed to simulate common workspace conditions encountered by visually impaired and neurodivergent users. To emulate the limited motion typical of visually impaired users who may have restricted head or body movement, video capture was conducted under multiple static perspectives, including front-facing view (simulating forward gaze), 180-degree lateral sweeps (mimicking head turning), and full 360-degree rotations while standing still (comprehensive environmental scanning). 
Furthermore, the visual results of our proposed solution can be seen in \ref{figure3} (a).

\subsubsection{Large Architectural Building}
The large architectural building environment presented complex indoor navigation challenges with multiple rooms, lifts, stairs, varied ceiling heights, diverse furniture configurations, and architectural features including columns, doorways, and open spaces. Moreover, the current pre-trained SpatialLM does not have the capability of identifying stairs, lifts and more detailed indoor objects; however, in the future, we can always fine-tune the SpatialLM model and increase its potential to classify more objects and events related to visually impaired and neuro-divergent users. In this environment, we also follow the same set of settings to capture the video data like the laboratory one, and we collected 12 videos in different settings. The results from these videos can be seen in the Figure \ref{figure3}(b).

\subsection{Results In Terms of Bird Eye View}

\begin{figure*}[htbp]
\centering
\includegraphics[width=0.9\textwidth]{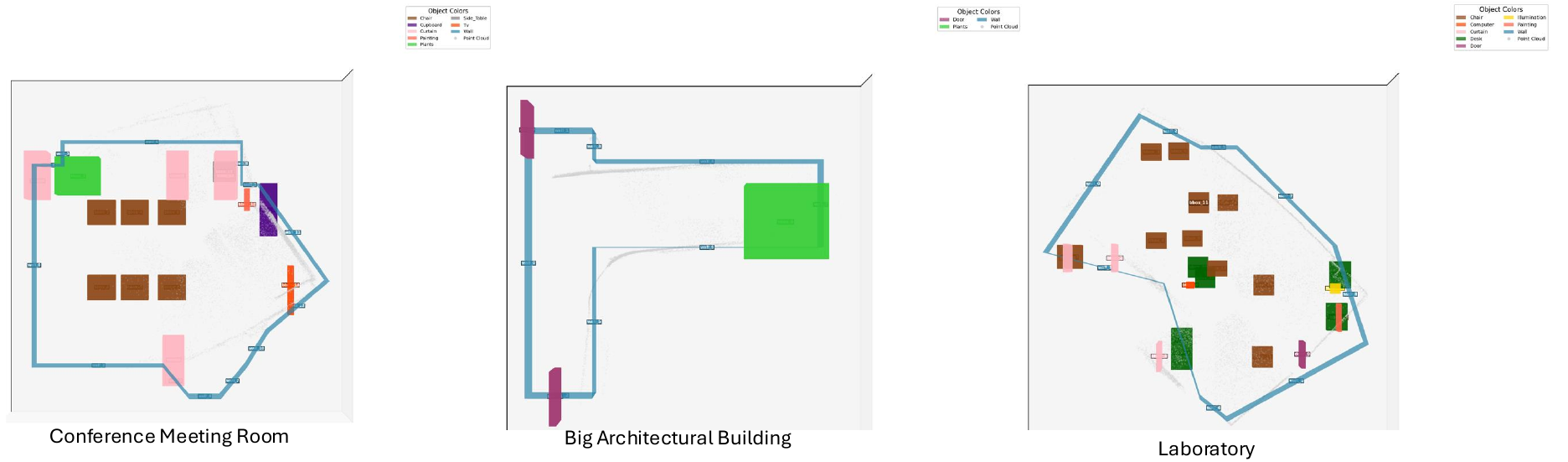} %
\caption{Figure shows the results in the form of BEV for selected indoor environments, especially how we can build basic layout of structures from our integrated system.}
\label{figure4}
\end{figure*}

The Bird's Eye View (BEV) visualisations presented in Figure \ref{figure4} provide an intuitive and comprehensive spatial overview of our system's detection and reconstruction capabilities across three distinct testing environments. 
In the laboratory environment Figure \ref{figure4}, the BEV reveals a densely packed workspace layout with multiple desks, chairs, computer equipment, and storage areas, demonstrating our system's precision in reconstructing constrained academic environments where navigation requires careful obstacle avoidance. The conference meeting room visualisation showcases a more open layout centered around large tables with fewer peripheral objects, which illustrates the system's versatility in handling diverse room configurations typical of professional meeting spaces. Most significantly, for the large architectural building BEV Figure \ref{figure4} captures complex multi-room layouts with interconnected spaces, corridors, and various furniture arrangements. 

\subsection{Results from Locally Deployed Open Source GPT-OSS-20B}

\begin{figure*}[htbp]
\centering
\includegraphics[width=0.99\textwidth]{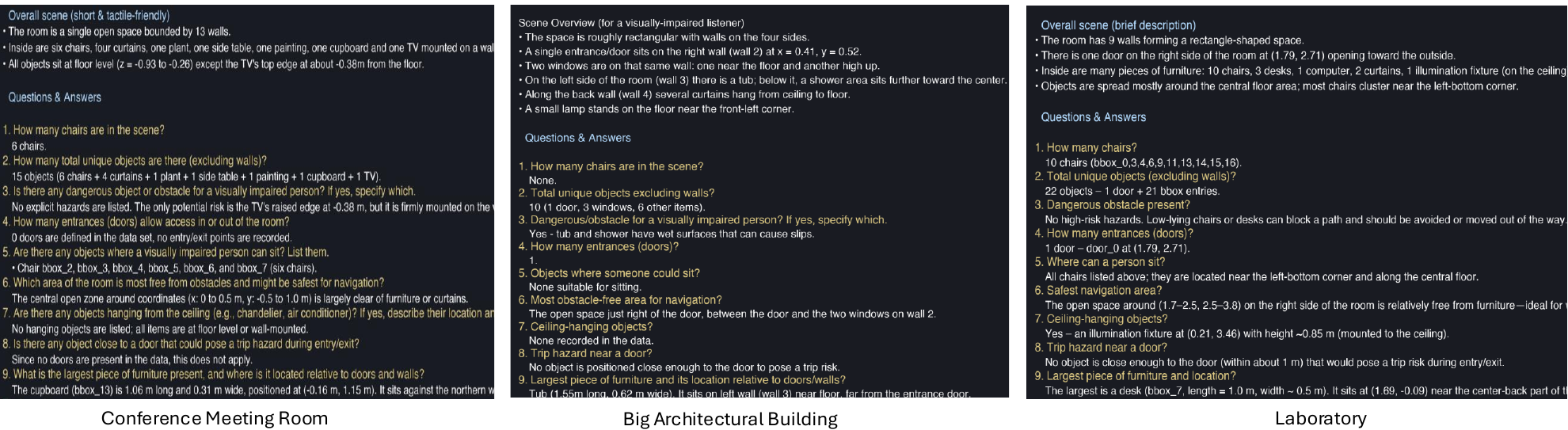} %
\caption{3D scene description and spatial reasoning responses generated by GPT-OSS-20B across three distinct testing environments: Conference Meeting Room, Big Architectural Building, and Laboratory, explain the ability of the proposed method to process SpatialLM's structured 3D scene data into detailed spatial descriptions. }
\label{figure5}
\end{figure*}

The experimental results demonstrate GPT-OSS-20B's exceptional capability to synthesise SpatialLM's structured 3D representations into contextually rich, navigation-focused descriptions that address the specific spatial reasoning needs of visually impaired and neurodivergent users. Across the three distinct testing environments presented in Figure \ref{figure5}, the system exhibits remarkable environmental adaptability and precision in spatial understanding. In the Conference Meeting Room environment, the system accurately identifies 6 chairs among 15 total objects while correctly noting the absence of entry/exit points, which is crucial information for spatial orientation in enclosed spaces. 

The model explains sophisticated safety awareness by identifying the TV's raised edge at -0.38m as a potential hazard while providing precise navigation coordinates for the central open zone (x: 0 to 0.5 m, y: -0.5 to 1.0 m). Conversely, in the Big Architectural Building setting, GPT-OSS-20B showcases critical safety assessment capabilities by correctly identifying wet surfaces from the tub (1.55m long, 0.62m wide) and shower as slip hazards, this information vital for preventing accidents in high-risk environments which can be seen in Figure \ref{figure5}. The Laboratory environment presents the most complex spatial arrangement with 22 objects, where the system successfully processes dense furniture clustering patterns, accurately locating the ceiling-mounted illumination fixture at specific coordinates (0.21, 3.46) with height specifications (~0.85m), and identifying optimal navigation corridors in the "open space around 1.7-2.5, 2.5-3.8" area.

The stark contrast between environments from the furniture-sparse Conference Meeting Room (15 objects) to the object-dense Laboratory (22 objects, 10 chairs), shows the system's scalability and adaptability. 
Particularly noteworthy is the model's consistent ability to provide actionable spatial guidance: identifying "no suitable seating" in the environment while providing detailed chair locations and bounding box references (bbox\_0,3,4,6,9,11,13,14,15,16) in the Laboratory setting shown in Figure \ref{figure5} 

Most significantly, the model's responses shows advanced spatial reasoning through detailed furniture positioning relative to walls and doors, such as identifying the largest furniture piece (desk: 1.0m x 0.5m) at coordinates (1.69, -0.09) in the Laboratory, positioned "roughly 2m from the door and 1.4m from the nearest wall." This level of granular spatial intelligence, combined with consistent safety prioritization and context-aware adaptation, validates the robustness of our integrated SLAM3R-SpatialLM-GPT-OSS-20B pipeline and shows its transformative potential for real-world assistive navigation applications serving visually impaired and neurodivergent populations.

\subsection{Unaligned Vs Aligned Results}

It is also important to report the results before and after the alignment. Therefore, Table \ref{tab:point_cloud_alignment} shows the effect of our proposed alignment on the detection algorithm. It is clear that before the alignment, the 3D VLM is detecting less number of objects in all the environments with misaligned axes, as shown in figure \ref{figure2}, and after the alignment, our proposed system is detecting more objects. These results directly support our methodology's emphasis on preprocessing quality for enabling accurate 3D spatial intelligence.

\begin{table}[htbp]
\centering
\caption{Number of objects detected using SpatialLM before and after scaling and alignment}
\label{tab:point_cloud_alignment}
\begin{tabular}{|l|c|c|}
\hline
\textbf{Environment Names} & \textbf{Unaligned} & \textbf{Aligned} \\
\hline
Conference Room & 13 & 28 \\
\hline
Laboratory $180^\circ$ & 8 & 22 \\
\hline
Laboratory $360^\circ$ & 4 & 41 \\
\hline
Big Building $180^\circ$ & 10 & 11 \\
\hline
Big Building $360^\circ$ & 8 & 16 \\
\hline
\end{tabular}
\end{table}

\section{Conclusion \& Future Research Directions} \label{conclusion_future_research}

This paper presents a novel end-to-end assistive navigation system that successfully integrates SLAM3R-based 3D reconstruction, SpatialLM scene understanding, and GPT-OSS-20B NLP to provide comprehensive spatial intelligence for visually impaired and neurodivergent users. Our system displays the feasibility of transforming smartphone video input into 3D spatial descriptions through a seamless four-stage pipeline including, video frame extraction, point-cloud generation, point-cloud alignment \& scaling, running through the 3D-VLM model, and then finally using LLM for 3D description and spatial reasoning. This paper provides experiments across diverse indoor environments, including laboratory workspaces, conference facilities, and architectural buildings. 
Similarly, the integration of structured 3D scene representations with NLP capabilities addresses fundamental limitations of existing 2D-based assistive technologies, which offer users unprecedented spatial awareness and independence in complex indoor environments where traditional navigation support fails to provide adequate spatial understanding. This indicates a clear foundation for next-generation assistive navigation technologies that bridge the gap between visual perception and spatial cognition through cutting-edge AI integration.

Future work will focus on expanding the capabilities of the system through fine-tuning SpatialLM to recognise additional object categories specifically relevant to assistive navigation, including stairs, elevators, accessibility ramps, and specialised institutional signage commonly encountered in hospitals, universities, and transportation hubs. 
Furthermore, the detection of crowded places and accidents could also be added to the label space for neurodivergent users. 
Similarly, for the evaluation, the future work will also contain an annotation benchmark that can verify, compare, and improve results from such integration of technologies.

\begin{credits}
\subsubsection{\ackname} This publication has emanated from research conducted with the financial support of Taighde Éireann – Research Ireland  [12/RC/2289\_P2] at Insight Research Ireland Centre for Data Analytics, Dublin City University.

\subsubsection{\discintname}
The authors have no competing interests to declare that are relevant to the content of this article.
\end{credits}

%
%
\bibliographystyle{splncs04}
\bibliography{references}

\end{document}